\def\year{2021}\relax
\documentclass[letterpaper]{article} 
\usepackage{aaai21}  
\usepackage{times}  
\usepackage{helvet} 
\usepackage{courier}  
\usepackage[hyphens]{url}  
\usepackage{graphicx} 
\usepackage{natbib}  
\usepackage{caption} 
\usepackage{microtype}
\usepackage{cite}
\usepackage{amsmath,amssymb,amsthm}
\usepackage{amsfonts}

\usepackage{enumitem} 

\usepackage{subcaption} 

\usepackage{multirow}

\usepackage{algorithm}
\usepackage{algpseudocode}
\algrenewcommand\algorithmicrequire{\textbf{Input:}}
\algrenewcommand\algorithmicensure{\textbf{Output:}}
\usepackage{tabularx}
\makeatletter
\newcommand{\multiline}[1]{%
  \begin{tabularx}{\dimexpr\linewidth-\ALG@thistlm}[t]{@{}X@{}}
    #1
  \end{tabularx}
}
\makeatother

\def\onedot{\ifx\@let@token.\else.\null\fi\xspace}

\makeatother

\def\Vec#1{{\boldsymbol{#1}}} 
\def\Mat#1{{\boldsymbol{#1}}} 
 
\title{Semi-Supervised Metric Learning: A Deep Resurrection}

\author{
    Ujjal Kr Dutta,\textsuperscript{1,2} Mehrtash Harandi,\textsuperscript{3} C Chandra Sekhar\textsuperscript{2}\\ 
}
\affiliations{
    \textsuperscript{1} Data Sciences, Myntra, India \\
    \textsuperscript{2}Dept. of Computer Science and Eng., Indian Institute of Technology Madras, India\\
    \textsuperscript{3}Dept. of Electrical and Computer Systems Eng., Monash University, Australia\\
    {ukd@cse.iitm.ac.in, ujjal.dutta@myntra.com, mehrtash.harandi@monash.edu, chandra@cse.iitm.ac.in}
}

\begin{document}

\maketitle

\begin{abstract}
Distance Metric Learning (DML) seeks to learn a discriminative embedding where similar examples are closer, and dissimilar examples are apart. In this paper, we address the problem of Semi-Supervised DML (SSDML) that tries to learn a metric using a few labeled examples, and abundantly available unlabeled examples. SSDML is important because it is infeasible to manually annotate all the examples present in a large dataset. Surprisingly, with the exception of a few classical approaches that learn a linear Mahalanobis metric, SSDML has not been studied in the recent years, and lacks approaches in the deep SSDML scenario. In this paper, we address this challenging problem, and revamp SSDML with respect to deep learning. In particular, we propose a stochastic, graph-based approach that first propagates the affinities between the pairs of examples from labeled data, to that of the unlabeled pairs. The propagated affinities are used to mine triplet based constraints for metric learning. We impose orthogonality constraint on the metric parameters, as it leads to a better performance by avoiding a model collapse.
\end{abstract}

\section{Introduction and Motivation}
Distance Metric Learning (DML) seeks to learn a discriminative embedding where similar examples are closer, and dissimilar examples are apart. The importance of DML is evidenced by the plethora of recent approaches introduced at the premier conferences in Computer Vision (CV) \cite{tuplet_margin_ICCV19,multi_sim_CVPR19,SNR_CVPR19,circle_CVPR20,fastAP_CVPR19,proxyNCA_ICCV17,arcface_CVPR19,soft_triple_ICCV19}, as well as Artificial Intelligence (AI) \cite{SUML_AAAI20,gu2020symmetrical,chen2020compressed,li2020symmetric,gong2020online} in general. In contrast to \textit{classification losses} that learn a class logit vector, \textit{embedding losses} in DML capture the relationships among examples. This makes embedding losses more general in nature, because of their flexibility to provide supervisory signals in the form of pairs \cite{contrastive}, triplets \cite{FaceNet,LMNN}, tuples \cite{N_pair} etc.
\begin{figure*}[t]
\centering
\begin{subfigure}{0.61\linewidth}
    	\centering
		\includegraphics[width=\linewidth]{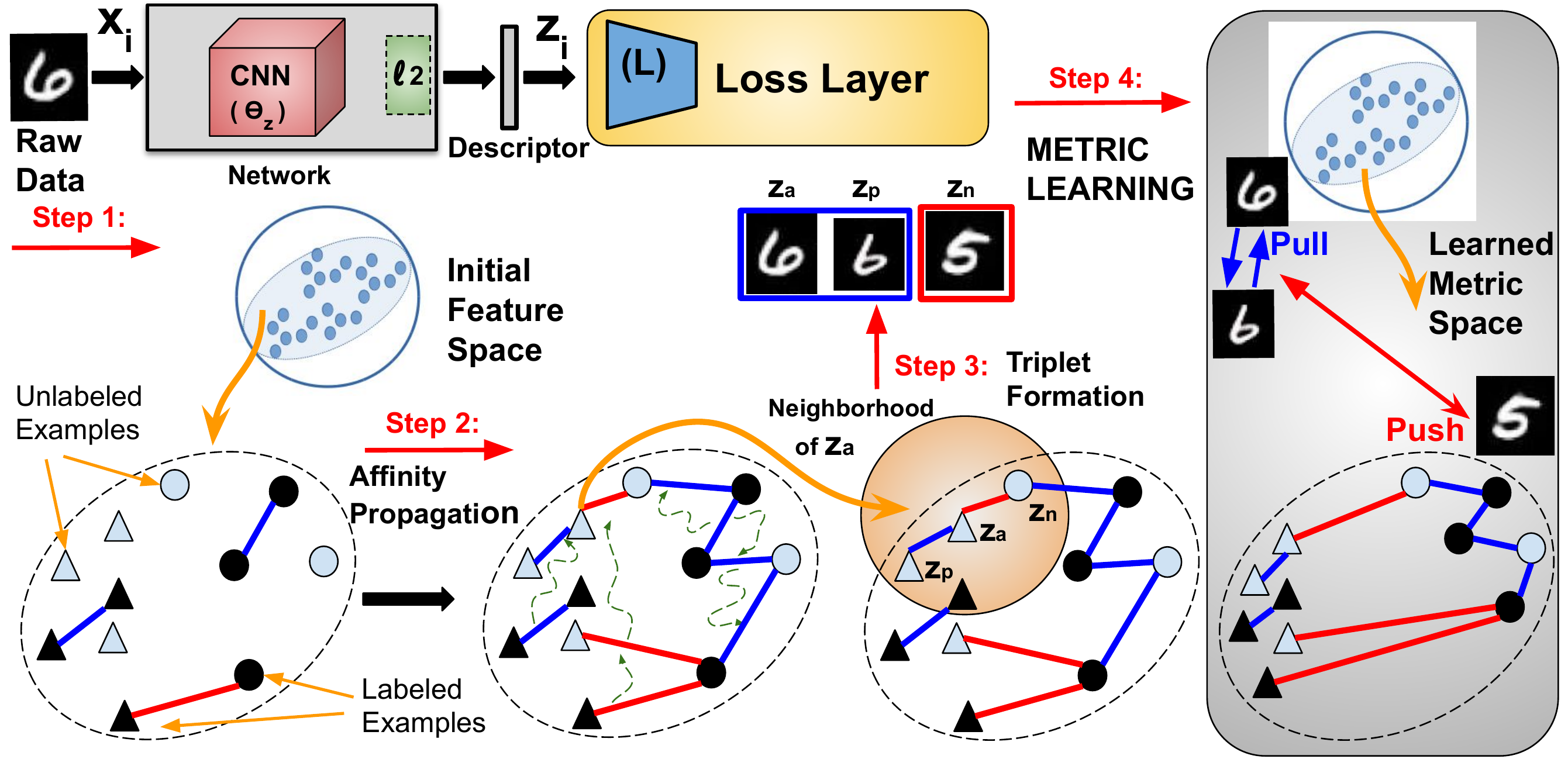}
		\caption{}
        \label{framework_agml}
\end{subfigure}
\hspace{0.2cm}
\begin{subfigure}{0.34\linewidth}
    	\centering
		\includegraphics[width=\linewidth]{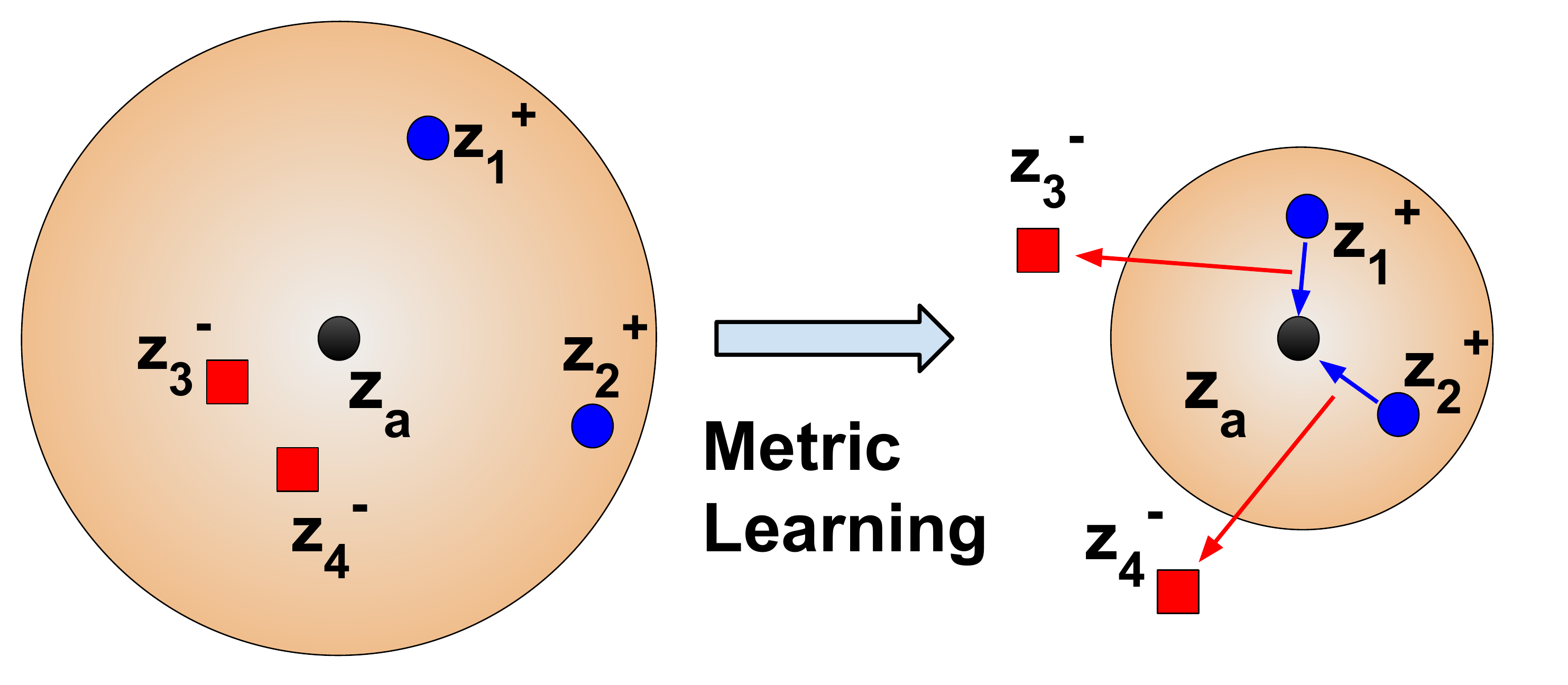}
		\caption{}
        \label{neigh_trip_mine}
\end{subfigure}
\caption{(Best viewed in color) (a) Illustration of our method. The raw image belongs to the MNIST dataset \cite{MNIST}, (b) Triplet mining around an anchor $\Vec{z}_a$ (shown in black), within its $k$- neighborhood $\mathcal{N}_k(\Vec{z}_a)$, $k$= 4. Points in blue ($\Vec{z}_1^+,\Vec{z}_2^+$) are more \textit{semantically} similar (by virtue of \textit{propagated affinities}) to the anchor, than the points in red ($\Vec{z}_3^-,\Vec{z}_4^-$). Hence, they should be pulled closer to the anchor in the learned space, compared to the red ones. }
\label{framework_mining}
\vspace{-0.6cm}
\end{figure*}

Despite the presence of many exemplary, state-of-the-art DML approaches as mentioned above, they are \textit{supervised} in nature. In order to provide constraints (pairs/ triplets/ tuples) for metric learning, they require manual annotations (\textit{class labels}). For large scale datasets, it is infeasible to obtain class labels for all the examples. In practical scenarios where it is possible to have a few examples annotated, with an abundance of additional unlabeled data, \textit{semi-supervised learning} \cite{oliver2018realistic,SSL_05,SSL_09,GAM_SSL_NeurIPS19,VAT_PAMI18,iscen2019label} can be applied. Surprisingly, with the exception of a few classical approaches that learn a linear Mahalanobis metric \cite{S3ML,LRML-b,SERAPH-b,ISDML,APSSML_ICANN18,Distributed_SERAPH_Access16}, the problem of Semi-Supervised DML (SSDML) has not been studied in the recent years, not to mention the lack of SSDML approaches in the deep DML scenario. In this paper, we address this challenging problem, and revamp SSDML with respect to deep learning. To the best of our knowledge, deep SSDML has been studied for the first time (this is despite the presence of deep semi-supervised approaches for classification losses).

As with any DML technique, our approach covers the following \textbf{two major aspects of DML}: 1. \textbf{Constraint Mining:} To appropriately mine constraints of examples (eg., pairs or triplets), and 2. \textbf{DML Loss:} An appropriate loss formulation to learn a metric using the mined constraints. In the recent years, a huge number of \textit{supervised, state-of-the-art} DML techniques have been proposed: Tuplet Margin \cite{tuplet_margin_ICCV19}, Multi-Similarity \cite{multi_sim_CVPR19}, SNR \cite{SNR_CVPR19}, Circle Loss \cite{circle_CVPR20}, FastAP \cite{fastAP_CVPR19}, ArcFace \cite{arcface_CVPR19}, Soft Triple \cite{soft_triple_ICCV19} etc. All of these approaches have made contributions either in terms of constraint mining, or a novel loss.

However, recently \cite{DML_reality_check_ECCV20} and \cite{revisiting_DML_strategies_ICML20} observed that when ran under the exact same experimental protocol (network architecture, embedding size, optimizers etc), classical metric learning losses involving \textit{pairs} \cite{contrastive} or \textit{triplets} \cite{FaceNet,LMNN} perform at par (and at times better) with that of the \textit{state-of-the-art} loss functions. For this reason, we make use of a classical \textit{triplet based} approach (because of the fact that the \textit{pairwise} loss still has the theoretical downside of applying the same distance threshold to all the pairs, irrespective of variances in the similarities or dissimilarities). In particular, we make use of an angular variant of a triplet-based metric learning loss, due to certain well-established theoretical benefits \cite{Angular_loss} (eg., ease of tuning an angle as compared to a distance hyperparameter, capture of third-order information by the angular variant, scale and rotation invariance, and so on).

Following are the \textbf{major contributions} of the paper:
\begin{itemize}
    \item Though end-to-end training is widely studied in supervised DML, the same is not the case for SSDML (a few classical SSDML approaches have attempted to learn a linear Mahalanobis metric, but not in an end-to-end fashion). \textbf{We revisit key linear SSDML baselines, and discuss their deep extensions}. Additionally, to address the theoretical limitations of prior SSDML methods, we also propose a \textbf{novel approach consisting of \textit{well-studied components}}: i) Affinity propagation \cite{S3ML,APSSML_ICANN18} to propagate information from the limited labeled examples to the unlabeled data using a graph, and ii) Making use of an \textit{adapted} angular variant of a triplet-based metric learning loss \cite{Angular_loss}. It should be noted that we formulate a simple approach with well-known components, and the superiority and \textit{novelty} of our approach lies by the \textbf{\textit{design composition}}, which we establish both empirically, as well as qualitatively. 
    \item \textbf{Constraint mining} and scalability: It should be noted that having propagated the affinities, we propose a \textit{novel triplet mining strategy} by considering the neighborhood of an example. Also, to make the approach scalable to large datasets, our approach is \textit{stochastic} in nature (absent in earlier classical SSDML methods).
    \item \textbf{Addressing model collapse}: A \textit{model collapse} could occur in DML which can project representations of close, but distinct examples to a single point (degenerate embeddings). To address this, we impose orthogonality on the metric and establish that it can lead to better embeddings. Arguably, some supervised works \cite{MDMLCLDD} have studied orthogonality while trying to alleviate class imbalance, achieving compact bases, etc. But our context is different. We explicitly show that it alleviates a collapse. This is particularly important in the semi-supervised case where we do not have much supervision, and a model may easily lead to degenerate embeddings.
\end{itemize}
An illustration of our method is shown in Figure \ref{framework_agml}.

\section{Extending Classical SSDML to Deep SSDML}
We begin our paper with the introduction of a few representative SSDML methods, while discussing their main principles. Let $\mathcal{X}=\mathcal{X}_L \cup \mathcal{X}_U$ be a given dataset, consisting of a set of labeled examples $\mathcal{X}_L=\{\Vec{z}_i\in \mathbb{R}^d \}_{i=1}^{N_L}$ with the associated label set $\{y_i\}_{i=1}^{N_L}, (y_i\in\{1,\cdots,C\}; C \textrm{ is the number of classes})$, and a set of unlabeled examples $\mathcal{X}_U=\{\Vec{z}_j \in \mathbb{R}^d\}_{j=N_L+1}^{N}$. Existing SSDML approaches learn the parametric matrix $\Mat{M}\in \mathbb{R}^{d \times d}, \Mat{M} \succeq 0$ of the squared Mahalanobis distance metric $\delta^2_{\Mat{M}}(\Vec{z}_i,\Vec{z}_j)=(\Vec{z}_i-\Vec{z}_j)^\top \Mat{M} (\Vec{z}_i-\Vec{z}_j)$, for a pair of descriptors $\Vec{z}_i,\Vec{z}_j \in \mathbb{R}^d$ (classically obtained using hand-crafted features). The SSDML approaches can mainly be categorized under two major paradigms: i) entropy minimization \cite{entropy_minimization_NIPS05}, and ii) graph-based. The SERAPH \cite{SERAPH-b,Distributed_SERAPH_Access16} approach is the only SSDML representative from the first class, and is expressed as \cite{SERAPH-b}:
\begin{equation}
\label{SERAPH_eqn}
{\small
\begin{aligned}
&\min_{\Mat{M}} -[\sum_{\substack{(\Vec{z}_i,\Vec{z}_j): \Vec{z}_i,\Vec{z}_j \in \mathcal{X}_L\\ y_{ij}=-1, y_i \neq y_j \\ y_{ij}=1, y_i = y_j}} \textrm{log }p_{ij}^{\Mat{M}}(y_{ij}) + \\& \mu \sum_{ \substack{(\Vec{z}_i,\Vec{z}_j): \\ \Vec{z}_i,\Vec{z}_j \in \mathcal{X}_U} } \sum_{y\in \{-1,1\}} p_{ij}^{\Mat{M}}(y) \textrm{log }p_{ij}^{\Mat{M}}(y)]+\lambda \textrm{Tr}(\Mat{M}).
\end{aligned}
}
\end{equation}
The first term maximizes the entropy of a conditional probability distribution over the pairwise constraints obtained from labeled data, while the second term minimizes the entropy  over the unlabeled data. Here, $p_{ij}^{\Mat{M}}(y_{ij})$ (and correspondingly $p_{ij}^{\Mat{M}}(y)$) denotes a probability parameterized by the distance metric $\delta^2_{\Mat{M}}(\Vec{z}_i,\Vec{z}_j)$, as follows: $p_{ij}^{\Mat{M}}(y_{ij})=\frac{1}{1+\textrm{exp}(y_{ij}(\delta^2_{\Mat{M}}(\Vec{z}_i,\Vec{z}_j)-\eta))}$. The regularizer $\textrm{Tr}(\Mat{M})$ ensures a \textit{projection sparsity} \cite{SERAPH-b}. $\eta,\mu,\lambda>0$ are hyperparameters in (\ref{SERAPH_eqn}).

The other category of SSDML approaches, i.e., graph-based ones, has majority of approaches to its disposal. Here we discuss the most prominent ones. The LRML \cite{LRML-b} method is expressed as:
\begin{equation}
\label{LRML_eqn}
{\small
\begin{aligned}
&\min_{\Mat{M}\succeq 0, \text{log }
|\Mat{M}|\geq 0}  \gamma_S \sum_{\substack{(\Vec{z}_i,\Vec{z}_j): \Vec{z}_i,\Vec{z}_j \in \mathcal{X}_L \\ y_i = y_j}} \delta^2_{\Mat{M}}(\Vec{z}_i,\Vec{z}_j)- \\& \gamma_D \sum_{\substack{(\Vec{z}_i,\Vec{z}_j): \Vec{z}_i,\Vec{z}_j \in \mathcal{X}_L\\ y_i \neq y_j }} \delta^2_{\Mat{M}}(\Vec{z}_i,\Vec{z}_j) + \textrm{Tr}(\Mat{M}\Mat{X}\mathcal{L}\Mat{X}^\top).
\end{aligned}
}
\end{equation}
The first and second terms in (\ref{LRML_eqn}) \textit{minimize} and \textit{maximize} the pairwise distances among similar and dissimilar pairs from labeled data, respectively. The third term is a \textit{Laplacian regularizer} that preserves the pairwise distances among all the examples in $\mathcal{X}$ (and hence, unlabeled data as well). Here, columns of $\Mat{X}\in \mathbb{R}^{d \times N}$ represent the descriptors of elements of $\mathcal{X}$, and $\mathcal{L}$ is the \emph{graph Laplacian} matrix obtained using the affinity matrix of the underlying kNN graph used.

The recently proposed graph-based ISDML \cite{ISDML} approach is a state-of-the-art among SSDML methods. Similar to LRML, it seeks to optimize the pairwise distances between labeled examples. However, it also takes into account the density around an unlabeled example while computing the Laplacian regularizer. The APLLR and APIT methods have also been proposed under a common framework \cite{APSSML_ICANN18}, that instead of performing naive optimization of the distances among labeled pairs, first computes a prior metric using the labeled data. The APLLR method computes the prior metric by optimizing a log-likelihood ratio, while the APIT method computes the prior metric using an information-theoretic approach. The final metric is learned using affinity propagation \cite{S3ML} on unlabeled data while staying close to the prior one.

\textbf{Extending the above methods for end-to-end learning:}
\begin{algorithm}[t]
\begin{footnotesize}
\caption{Stochastic extension of the SSDML baselines}
\label{alg_AGML}
\begin{algorithmic}[1]
\State Initialize $\theta_z^0, \Mat{M}_0$.
\For {$t \gets 1$ to $T$}
\State \multiline{Fix $\theta_z^{t-1}$ and learn $\Mat{M}_t$ using the given SSDML method. \label{update_M}}
\State \multiline{Fix $\Mat{M}_t$ and learn $\theta_z^t$ by backpropagation using SGD. \label{update_th}}
\EndFor\label{for_count}
\State \textbf{return} $\theta_z^T,\Mat{M}_T$
\end{algorithmic}
\end{footnotesize}
\end{algorithm}
Although trivial, the methods discussed above have not been studied in the context of end-to-end deep learning. It is noteworthy that as the losses in these methods are differentiable, we can backpropagate their gradients while using mini-batch based stochastic optimization. Due to the graph-based nature of LRML, ISDML, APLLR and APIT, one could first sample a random partition of the unlabeled examples, and construct a sub-graph along with the limited number of available labeled examples. Then, the mini-batches can be drawn from this partition alone for a number of epochs (over this partition). This process could be iterated over a number of partitions over the entire dataset. The same partition based strategy could be used for SERAPH as well, with the only exception that it does not require graph construction.

Formally, we can learn a deep embedding that induces a distance metric of the form $\delta^2_{\Mat{M}}(\Vec{z}_i,\Vec{z}_j)$, such that the descriptors $\Vec{z}_i,\Vec{z}_j \in \mathbb{R}^d$ are obtained using a deep neural network $z:\mathcal{X}\rightarrow \mathbb{R}^d$ with parameters $\theta_z$, while simultaneously learning $(\theta_z,\Mat{M})$ in an end-to-end manner. This general stochastic approach is illustrated in Algorithm \ref{alg_AGML}.

\textbf{Limitations with existing SSDML methods:} Having proposed stochastic extensions for the existing classical SSDML techniques LRML, SERAPH, ISDML, APLLR and APIT, we now discuss their limitations. A fundamental weakness of the LRML method is the use of a naive Laplacian regularizer as: $\textrm{Tr}(\Mat{M}\Mat{X}\mathcal{L}\Mat{X}^\top) = \frac{1}{2}\sum_{i,j=1}^N\delta^2_{\Mat{M}}(\Vec{z}_i,\Vec{z}_j) W_{ij} $. Here, $W_{ij}$ refers to the affinity between the $i^{th}$ and $j^{th}$ examples, and is computed directly using the distances among the initial representations. Hence, the distance $\delta^2_{\Mat{M}}(\Vec{z}_i,\Vec{z}_j)$ in the learned space could be affected by a poor initial $W_{ij}$. The ISDML method scales the affinity terms by considering the densities around each individual example. However, both these techniques are prone to be affected by a poor initial representation. Moreover, they fail to adapt their affinities by the additionally present labeled data information.

The APLLR and APIT techniques make use of an affinity propagation principle \cite{S3ML} to propagate the information from the labeled pairs to the unlabeled ones. This leads to a more informative affinity matrix. However, their dependency on a prior pre-computed metric sometimes have adverse effects. Especially, in scenarios where there is an extremely limited amount of labeled data, the prior parametric matrix is often close to singular, and leads to a poor metric. Furthermore, despite the affinity propagation, they do not \textit{mine} informative constraints for metric learning (which is very crucial). This fails to fully leverage the benefits of affinity propagation. On the other hand, the SERAPH method is based on the \textit{entropy minimization} principle. Although it is a good principle due to its capability of minimizing the class overlap, the SERAPH method takes a long time to converge, and often gets trapped in poor local minima.

To address these limitations with the existing SSDML methods, we propose a new method. However, we propose that instead of learning the metric wrt the matrix $\Mat{M}$ ($O(d^2)$ parameters), we utilize the property that $\Mat{M} \succeq 0$, and factorize it as: $\Mat{M}=\Mat{L}\Mat{L}^\top$ s.t. $\Mat{L} \in \mathbb{R}^{d \times l}, l \leq d$, and learn the metric wrt $\Mat{L}$ with lesser parameters ($O(dl)$). The pseudo-code of our method is same as Algorithm \ref{alg_AGML} (after replacing $\Mat{M}$ with $\Mat{L}$). We now discuss our proposed method.

\section{Proposed Approach}

\subsection{Triplet Mining using Affinity Propagation}
In the semi-supervised DML setting, the first task is to \textit{mine} informative constraints, and then use an appropriate loss function. As the first stage, we propose a novel method for mining constraints using a graph-based technique while leveraging affinity propagation \cite{S3ML}. Let $\mathcal{X}_U^{(p)}$ be a randomly selected partition of unlabeled data. We construct a kNN graph using $\mathcal{X}_L \cup \mathcal{X}_U^{(p)}$, s.t. the nodes represent the examples, and edge weights denote the \textit{affinities} (or similarities) among the examples. The initial affinity matrix $\Mat{W}^0\in\mathbb{R}^{(N_L+N_p)\times (N_L+N_p)}$ is defined as follows: i) $W_{ij}^0 = +1 \mbox{, if } i \neq j \textrm{ and } \exists y_i, y_j, \textrm{ s.t. } y_i = y_j$, ii) $W_{ij}^0 = -1 \mbox{, if } i \neq j \textrm{ and } \exists y_i, y_j, \textrm{ s.t. } y_i \neq y_j$, iii) $W_{ij}^0 = +1 \mbox{, if } i = j$, and iv) $W_{ij}^0 =0 \mbox{} \textrm{, otherwise}$. Here, $N_L$ and $N_p$ are respective cardinalities of $\mathcal{X}_L$ and $\mathcal{X}_U^{(p)}$.

The neighborhood structure of the kNN graph can be indicated using a matrix $\Mat{Q} \in\mathbb{R}^{(N_L+N_p)\times (N_L+N_p)}$ defined as: i) $Q_{ij} = 1/k \mbox{, if } \Vec{z}_j \in \mathcal{N}_k(\Vec{z}_i)$, and ii) $Q_{ij} = 0 \mbox{} \textrm{, otherwise}$. Here, $\mathcal{N}_k(\Vec{z}_i)$ is the set of $k$-nearest neighbor examples of $\Vec{z}_i$. The flow of information from the non-zero entries of $\Mat{W}^0$ (representing the labeled pairs) to the zero entries (representing the unlabeled pairs), can be performed by using a closed-form expression described by Liu \textit{et al.} \cite{S3ML}, as: $\Mat{W}^*=(1-\gamma)(\mathbf{I}_{N_L+N_p}-\gamma \Mat{Q})^{-1}\Mat{W}^0$. This step is called \textit{affinity propagation}. This step essentially performs a random-walk to propagate the affinities, where $0<\gamma<1$ is a weight hyper-parameter between the affinities obtained at the current step to that of the initial one. As $N_L+N_p\ll N$, we do not encounter any difficulty while scaling up to large datasets. To obtain a symmetric affinity matrix $\Mat{W}$, we perform a final symmetrization step as follows: $W_{ij}=(W^*_{ij}+W^*_{ji})/2$.

The finally obtained representation of the symmetric affinity matrix $\Mat{W}$ is used to mine triplets for metric learning. In doing so, we take into account the $k$-neighborhood $\mathcal{N}_k(\Vec{z}_a)$ of an example $\Vec{z}_a\in \mathcal{X}_L \cup \mathcal{X}_U^{(p)}$ that we consider as an anchor (Figure \ref{neigh_trip_mine}). Let $\mathcal{N}_W(\Vec{z}_a)=\{\Vec{z}^+_1,\Vec{z}^+_2,\cdots,\Vec{z}^+_{k/2},\Vec{z}^-_{k/2+1},\Vec{z}^-_{k/2+2},\cdots,\Vec{z}^-_{k}\}$ be the $k$-neighboring examples of $\Vec{z}_a$ sorted in descending order of their \textit{propagated affinities} w.r.t. $\Vec{z}_a$, i.e., $\Mat{W}(\Vec{z}_a,\Vec{z}^+_1)>\cdots>\Mat{W}(\Vec{z}_a,\Vec{z}^+_{k/2})>\Mat{W}(\Vec{z}_a,\Vec{z}^-_{k/2+1})>\cdots>\Mat{W}(\Vec{z}_a,\Vec{z}^-_k)$. Essentially, $\mathcal{N}_W(\Vec{z}_a)$ is simply a sorted version of $\mathcal{N}_k(\Vec{z}_a)$. As the obtained affinities are an indication of the semantic similarities among examples, we take it as a guidance to form triplets. Given an anchor $\Vec{z}_a$, intuitively we can consider an example $\Vec{z}^+_i$ with more affinity towards $\Vec{z}_a$ as a \textit{positive}, and another example $\Vec{z}^-_j$ with lesser affinity towards $\Vec{z}_a$ as a \textit{negative}, and form a triplet $(\Vec{z}_a,\Vec{z}^+_i,\Vec{z}^-_j)$. By considering first half of examples in the sorted neighborhood $\mathcal{N}_W(\Vec{z}_a)$ as \textit{positives}, and remaining half as \textit{negatives}, we can form the following triplets: $(\Vec{z}_a,\Vec{z}^+_1,\Vec{z}^-_{k/2+1}), (\Vec{z}_a,\Vec{z}^+_2,\Vec{z}^-_{k/2+2}),\cdots, (\Vec{z}_a,\Vec{z}^+_{k/2},\Vec{z}^-_{k})$. One may select the set of anchors from entire $\mathcal{X}_L \cup \mathcal{X}_U^{(p)}$, or by seeking the modes of the graph, without loss of generality.
\begin{table*}[t]
\centering

\resizebox{0.9\linewidth}{!}{%
\begin{tabular}{|c|ccccc|ccccc|ccccc|}
\hline
\textbf{Dataset} & \multicolumn{5}{c|}{\textbf{MNIST}}                                           & \multicolumn{5}{c|}{\textbf{Fashion-MNIST}}                                   & \multicolumn{5}{c|}{\textbf{CIFAR-10}}                                        \\ \hline
\textbf{Method}  & \textbf{NMI}  & \textbf{R@1}  & \textbf{R@2}  & \textbf{R@4}  & \textbf{R@8}  & \textbf{NMI}  & \textbf{R@1}  & \textbf{R@2}  & \textbf{R@4}  & \textbf{R@8}  & \textbf{NMI}  & \textbf{R@1}  & \textbf{R@2}  & \textbf{R@4}  & \textbf{R@8}  \\ \hline
Initial          & 17.4          & 86.5          & 92.4          & 95.9          & 97.6          & 32.6          & 71.9          & 82.0          & 89.5          & 94.8          & 20.4          & 38.2          & 54.9          & 71.0          & 83.6          \\ \hline \hline
Deep-LRML              & \textbf{48.3} & 88.7          & 93.2          & 95.9          & 97.6          & \textbf{53.2} & 75.3          & 83.9          & 90.2          & 95.0          & 16.0          & 40.2          & 56.3          & 71.5          & 83.7          \\
Deep-SERAPH            & 45.3          & 92.5          & 95.7          & 97.6          & 98.7          & 53.0          & 75.9          & 85.1          & 91.5          & 95.4          & 21.3          & 39.5          & 56.2          & 71.1          & 83.7          \\
Deep-ISDML             & 44.1          & 92.1          & 95.7          & 97.5          & 98.6          & 51.3          & 74.4          & 84.2          & 90.7          & 95.1          & 20.0          & 36.4          & 52.4          & 68.5          & 82.1          \\
Deep-APLLR             & 30.5          & 57.6          & 71.1          & 82.1          & 90.1          & 38.9          & 59.5          & 73.3          & 84.1          & 91.4          & 13.3          & 24.4          & 40.0          & 58.8          & 76.4          \\
Deep-APIT              & 31.7          & 87.6          & 92.9          & 96.0          & 97.9          & 37.3          & 69.4          & 80.1          & 88.5          & 94.1          & 11.4          & 26.0          & 41.4          & 59.6          & 76.4          \\ \hline
\textbf{Ours}    & 47.5          & \textbf{93.9} & \textbf{96.6} & \textbf{98.2} & \textbf{98.9} & 52.1          & \textbf{77.6} & \textbf{86.0} & \textbf{91.8} & \textbf{95.6} & \textbf{25.3} & \textbf{41.4} & \textbf{57.3} & \textbf{72.6} & \textbf{84.9} \\ \hline
\end{tabular}%
}
\caption{Comparison against state-of-the-art SSDML approaches on MNIST, Fashion-MNIST and CIFAR-10.}
\label{ssdml_all_sota}
\vspace{-0.3cm}
\end{table*}

\begin{table*}[t]
\centering

\resizebox{0.6\linewidth}{!}{%
\begin{tabular}{|c|ccccc|ccccc|}
\hline
\textbf{Dataset}     & \multicolumn{5}{c|}{\textbf{CUB-200} }                                             & \multicolumn{5}{c|}{\textbf{Cars-196} }                                            \\ \hline \hline
\textbf{Method}      & \textbf{NMI}  & \textbf{R@1}  & \textbf{R@2}  & \textbf{R@4}  & \textbf{R@8}  & \textbf{NMI}  & \textbf{R@1}  & \textbf{R@2}  & \textbf{R@4}  & \textbf{R@8}  \\ \hline
Initial              & 34.3          & 31.7          & 42.2          & 55.4          & 67.9          & 23.7          & 24.2          & 32.8          & 43.4          & 54.9          \\ \hline \hline
Deep-LRML                  & 49.9          & 34.9          & 45.0          & 55.4          & 65.7          & 40.5          & 33.2          & 41.2          & 49.4          & 58.3          \\
Deep-SERAPH                & 50.5          & 39.7          & 52.2          & 64.4          & 76.5          & 37.2          & 34.8          & 46.7          & 59.4          & 71.7          \\
Deep-ISDML                 & 50.6          & 43.7          & 55.7          & 67.6          & 78.4          & 25.8          & 30.6          & 40.8          & 52.2          & 63.2          \\
Deep-APLLR                 & 38.9          & 25.7          & 36.6          & 48.7          & 61.9          & 33.2          & 33.0          & 43.6          & 55.6          & 67.9          \\
Deep-APIT                  & 52.1          & 42.2          & 54.5          & 66.7          & 78.2          & 37.3          & 38.5          & 50.7          & 62.7          & 74.9          \\ \hline \hline
\textbf{Ours} & \textbf{54.0} & \textbf{44.8} & \textbf{56.9} & \textbf{69.1} & \textbf{79.9} & \textbf{40.7} & \textbf{45.7} & \textbf{58.1} & \textbf{69.5} & \textbf{80.4} \\ \hline
\end{tabular}%
}
\caption{Comparison against state-of-the-art SSDML approaches on fine-grained datasets.}
\label{results_all_cub_cars}
\vspace{-0.3cm}
\end{table*}

\begin{table*}[!ht]
\centering

\resizebox{0.6\linewidth}{!}{%
\begin{tabular}{|c|ccccc|ccccc|}
\hline
\textbf{Dataset}     & \multicolumn{5}{c|}{\textbf{CUB-200} }                                             & \multicolumn{5}{c|}{\textbf{Cars-196} }                                            \\ \hline \hline
\textbf{Method}      & \textbf{NMI}  & \textbf{R@1}  & \textbf{R@2}  & \textbf{R@4}  & \textbf{R@8}  & \textbf{NMI}  & \textbf{R@1}  & \textbf{R@2}  & \textbf{R@4}  & \textbf{R@8}  \\ \hline
    Exemplar   & 45.0          & 38.2          & 50.3          & 62.8          & 75.0  & 35.4          & 36.5          & 48.1          & 59.2          & 71.0          \\
    Rotation   &49.1	&42.5	&55.8	&68.6	&79.4   &32.7	&33.3	&44.6	&56.4	&68.5          \\
    NCE       & 45.1          & 39.2          & 51.4          & 63.7          & 75.8            & 35.6          & 37.5          & 48.7          & 59.8          & 71.5          \\
    DeepCluster & 53.0          & 42.9          & 54.1          & 65.6          & 76.2& 38.5          & 32.6          & 43.8          & 57.0          & 69.5          \\ 
    Synthetic &53.4 &43.5 &56.2 &68.3 &79.1  &37.6 &42.0 &54.3 &66.0 &77.2          \\ \hline
    \textbf{Ours}        & \textbf{54.0} & \textbf{44.8} & \textbf{56.9} & \textbf{69.1} & \textbf{79.9}        & \textbf{40.7} & \textbf{45.7} & \textbf{58.1} & \textbf{69.5} & \textbf{80.4} \\ \hline
\end{tabular}%
}
\caption{Comparison of our method against state-of-the-art deep unsupervised methods on fine-grained datasets.}
\label{cars_vs_unsup}
\vspace{-0.3cm}
\end{table*}

\subsection{Triplet-based Orthogonality-Promoting SSDML}
Given $\mathcal{X}_L \cup \mathcal{X}_U^{(p)}$ and the corresponding $\Mat{W}$, assume that we have obtained a triplet set $\mathcal{T}_p=\bigcup_b \mathcal{T}_p^{(b)}$. Here, $\mathcal{T}_p^{(b)}$ is a mini-batch of $T_b$ triplets, and $b \in [1, \cdots, \left \lfloor \frac{|\mathcal{T}_p|}{T_b}  \right \rfloor]$. Let, $\mathcal{T}_p^{(b)}=\{ (\Vec{z}_i,\Vec{z}_i^+,\Vec{z}_i^-) \}_{i=1}^{T_b}$. Then, we propose a smooth objective to learn the parameters ($\Mat{L},\theta_z$) of our deep metric as follows:
\begin{equation}
\label{opt_prob_AGML}
\begin{aligned}
\min_{\Mat{L},\theta_z} J_{metric}=\sum_{i=1}^{T_b}\textrm{log}(1+\textrm{exp}(m_{i})).
\end{aligned}
\end{equation}
Here, $m_{i}=\delta^2_{\Mat{L}}(\Vec{z}_i,\Vec{z}_i^+)-4\textrm{ tan}^2\alpha \textrm{ } \delta^2_{\Mat{L}}(\Vec{z}_i^-,\Vec{z}_{i-avg})$, s.t., $\Vec{z}_{i-avg}=(\Vec{z}_i+\Vec{z}_i^+)/2$, and $\delta^2_{\Mat{L}}(\Vec{z}_i,\Vec{z}_j)=(\Vec{z}_i-\Vec{z}_j)^\top\Mat{L}\Mat{L}^\top(\Vec{z}_i-\Vec{z}_j)$. $m_i$ tries to pull the anchor $\Vec{z}_i$ and the positive $\Vec{z}_i^+$ together, while moving away the negative $\Vec{z}_i^-$ from the mean $\Vec{z}_{i-avg}$, with respect to an angle $\alpha >0$ at the negative $\Vec{z}_i^-$ \cite{Angular_loss}. Given the fact that $J_{metric}$ in (\ref{opt_prob_AGML}) is fully-differentiable, we can backpropagate the gradients to learn the parameters $\theta_z$ using SGD. This helps us in integrating our method within an end-to-end deep framework using Algorithm \ref{alg_AGML} (except, we learn wrt $\Mat{L}$ instead of $\Mat{M}$).

Additionally, we constrain $\Mat{L}\in \mathbb{R}^{d \times l}, l \leq d$ to be a orthogonal matrix, i.e., $\Mat{L}^\top\Mat{L}=\mathbf{I}_l$. This is because in contrast to other regularizers that only constrain the values of the elements of a parametric matrix (like $l_1$ or $l_2$ regularizers), \textit{orthogonality} omits redundant correlations among the different dimensions, thereby omits redundancy in the metric which could hurt the generalization ability due to overfitting \cite{MDMLCLDD}. We further show that this avoids a \textit{model collapse} (degenerate embeddings of distinct, close examples collapsing to a singularity). Theoretically, this is by virtue of the Johnson–Lindenstrauss Lemma \cite{dasgupta2003elementary}, which ensures the following near-isometry property of the embedding: $(1-\epsilon)\left \| \Vec{z}_i - \Vec{z}_j \right \|_2^2 \leq \frac{d}{l} \left \| \Mat{L}^\top \Vec{z}_i - \Mat{L}^\top \Vec{z}_j \right \|_2^2 \leq (1+\epsilon)\left \| \Vec{z}_i - \Vec{z}_j \right \|_2^2$, $0<\epsilon<1$.

Given a mini-batch of triplets $\mathcal{T}_p^{(b)} =\{ (\Vec{z}_i,\Vec{z}_i^+,\Vec{z}_i^-) \}_{i=1}^{T_b}$, we provide an efficient matrix-based implementation to learn $\Mat{L}$ (assuming a fixed $\theta_z$), in the supplementary. Here, we present the computational time complexity of the major steps involved: 
\textbf{i) Cost Function:} Computing the cost requires four matrix multiplications resulting in complexity of O($ldT_b$). Next, the transpose of the matrix products need to be computed, requiring O($lT_b$). Finally, the sum of losses across all triplets can be computed using a matrix trace operation requiring O($T_b$) complexity. \textbf{ii) Gradients:} The gradient with respect to $\Mat{L}$ requires the following computations: transposes requiring O($dT_b$), outer products requiring O($d^2T_b$). The subsequent products require O($d^2l$). Hence, the overall complexity is O($dT_b+d^2T_b+d^2l$).

Our proposed algorithm is linear in terms of the number of triplets in a mini-batch, i.e., $T_b$, which is usually low. The complexity of our algorithm is either linear or quadratic in terms of the original dimensionality $d$ (which in practice is easily controllable within a neural network).

It should be noted that in contrast to existing SSDML approaches, our method also enjoys a lower theoretical computational complexity. For example, LRML has a complexity of O($d^3$), while APLLR and APIT are of the order $O(d^3+N^3)$.

\section{Experiments}

\begin{figure*}[t]
\centering
\minipage{0.30\textwidth}
  \includegraphics[width=\linewidth]{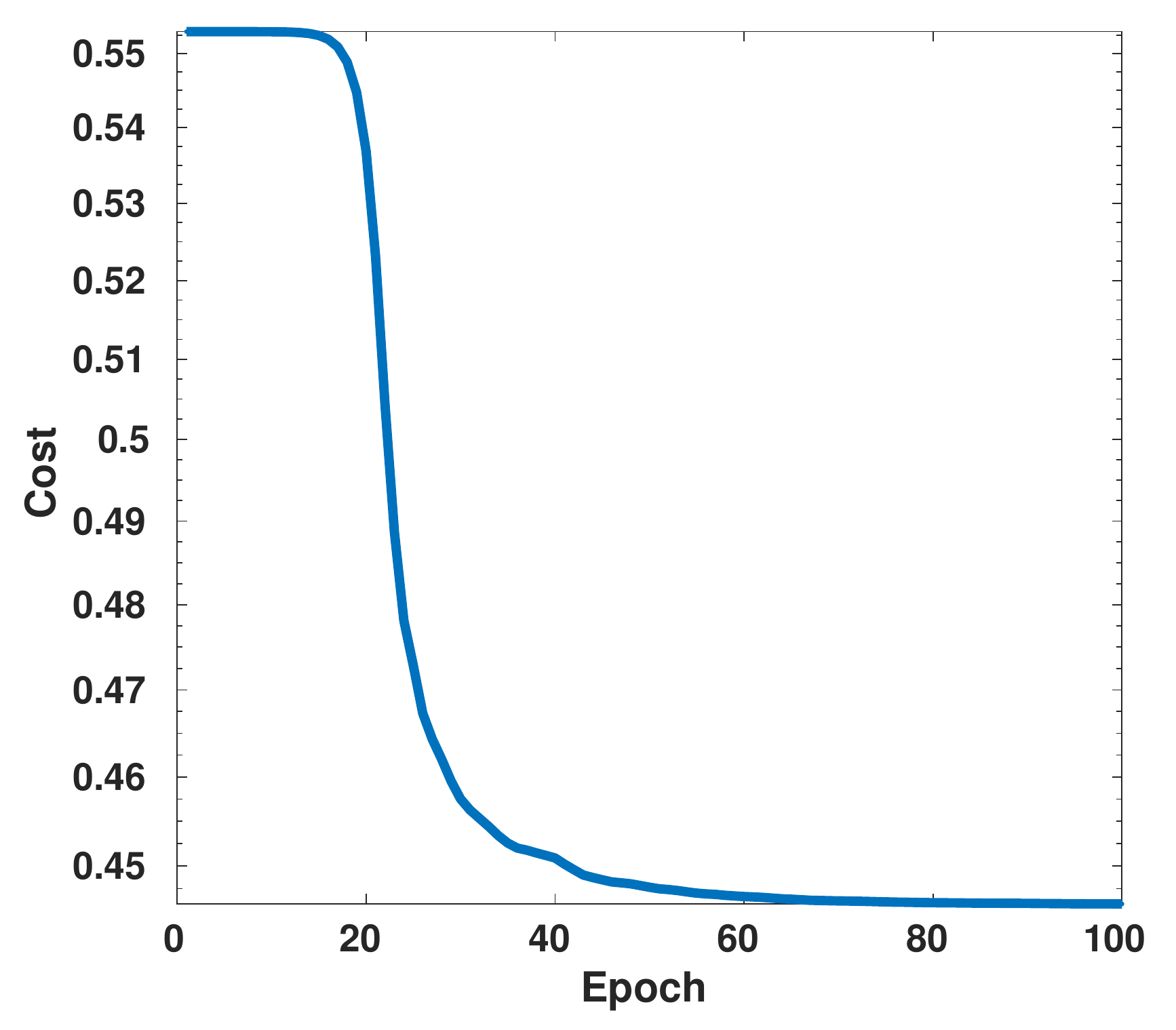}
\endminipage \hspace{0.1cm}
\minipage{0.30\textwidth}
  
    \resizebox{\linewidth}{!}{%
    \begin{tabular}{|c|ccccc|}
    \hline
    \textbf{$\alpha$} & \textbf{NMI} & \textbf{R@1} & \textbf{R@2} & \textbf{R@4} & \textbf{R@8} \\ \hline
    $35^\circ$             & 40.5         & 45.0         & 57.2         & 69.0         & 79.8         \\
    $40^\circ$             & 40.3         & 45.5         & 57.8         & 69.1         & 80.0         \\
    $45^\circ$             & 40.7         & 45.7         & 58.1         & 69.5         & 80.4         \\
    $55^\circ$             & 40.4         & 45.7         & 57.9         & 69.4         & 80.1         \\ \hline
    \end{tabular}%
    }
  
\endminipage \hspace{0.1cm}
\minipage{0.34\textwidth}%
  \includegraphics[width=\linewidth]{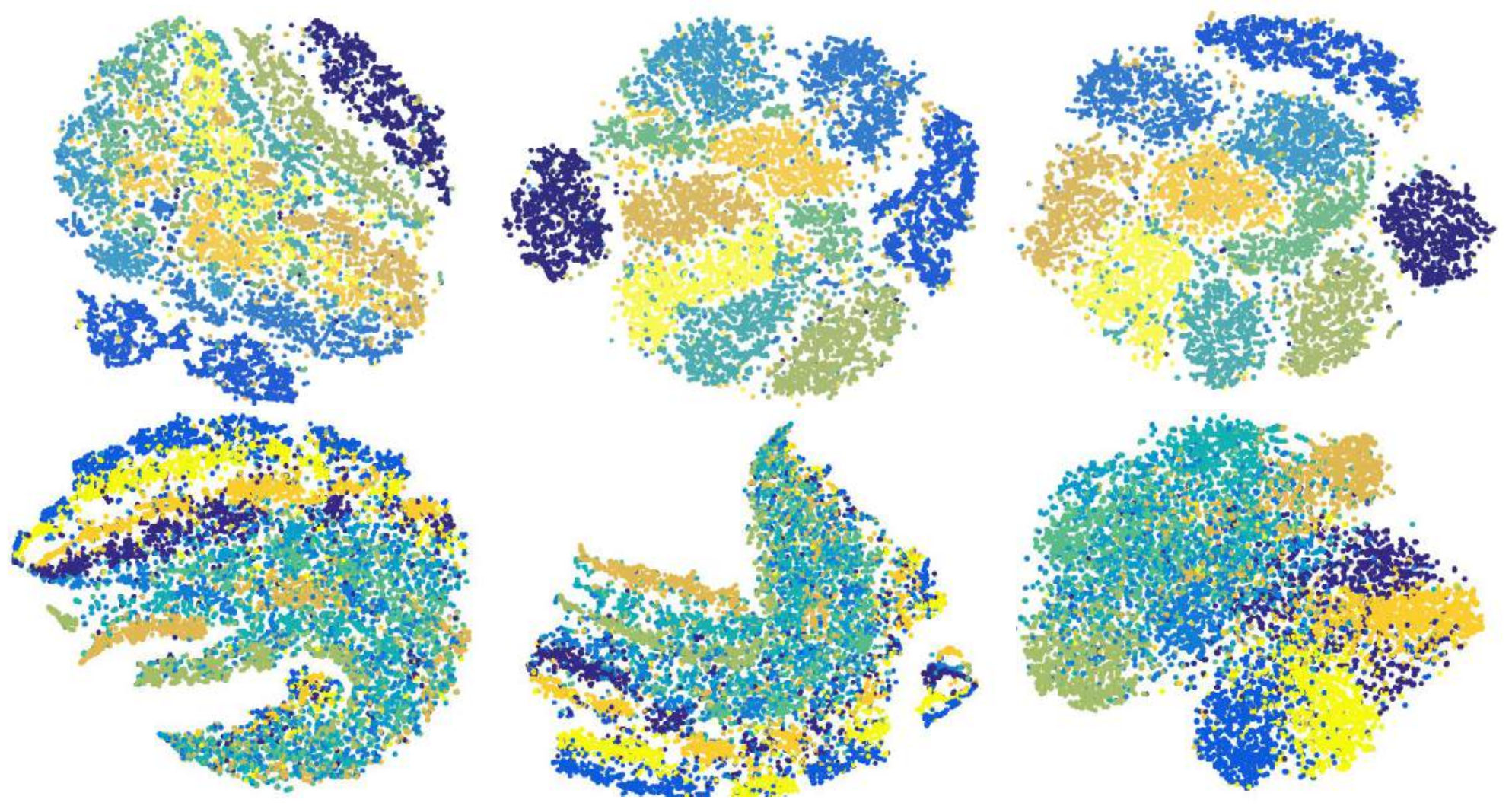}
\endminipage
\caption{Left: Convergence behaviour of our method on the Cars-196 dataset \cite{Cars196}. Middle: Ablation study showing sensitivity of our method towards $\alpha$ on Cars-196 \cite{Cars196} dataset. Right: tSNE embeddings for the test examples of MNIST (top row) and CIFAR-10 (bottom row). The left column represents the embeddings obtained right after random initialization. The embeddings obtained by our method: without orthogonality constraint on $\Mat{L}$ (middle column) and with orthogonality constraint on $\Mat{L}$ (rightmost column). Orthogonality leads to better embeddings (see Table \ref{ablation_quant_orthog}).}
\label{ablation_alpha}
\end{figure*}

\begin{figure*}[ht]
    \centering
    \includegraphics[width=0.8\linewidth]{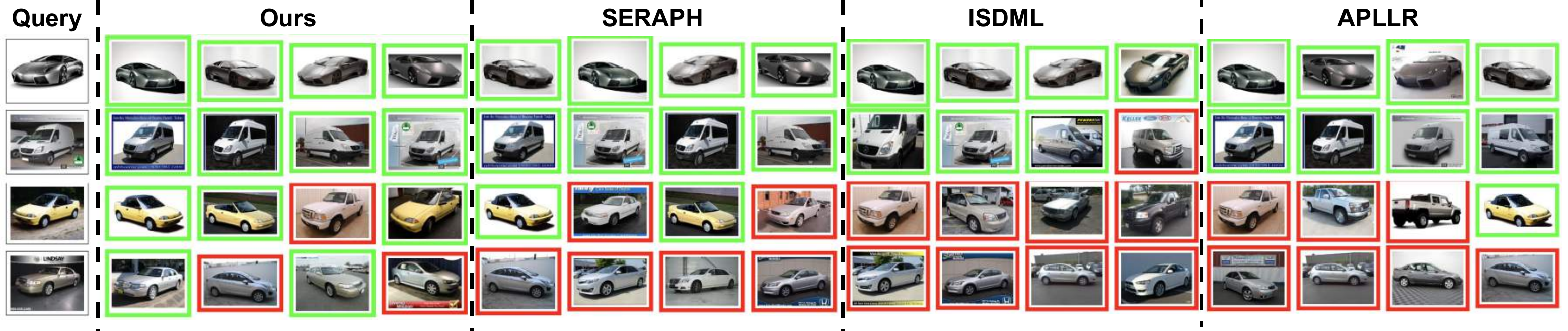}
    \caption{Qualitative comparison of retrieval performance against state-of-the-art SSDML approaches on the Cars-196 dataset. For a retrieved image, a red box denotes an incorrect retrieval (different class as the query), and a green box denotes a correct retrieval (same class as the query).}
    \label{quali_cars}
    \vspace{-0.6cm}
\end{figure*}

In this section we evaluate our proposed method in terms of its effectiveness in clustering and retrieval tasks on a number of benchmark datasets.

\textbf{Datasets:} Following recent literature, the benchmark datasets that have been used are as follows:
\begin{itemize}[noitemsep]
    \item \textbf{MNIST} \cite{MNIST}: It is a benchmark dataset that consists of 70000 gray-scale images of handwritten digits. Each image is of $28\times 28$ pixels. There are 60000 training images and 10000 test images in the standard split.
    \item \textbf{Fashion-MNIST} \cite{Fashion-MNIST}: It is a similar dataset as the MNIST, but consists of images from 10 categories of fashion products. There are 60000 training images and 10000 test images in the standard split.
    \item \textbf{CIFAR-10} \cite{CIFAR}: This dataset consists of colour images of $32 \times 32$ pixels, containing animal or vehicle objects from 10 different categories. There are 50000 training images and 10000 test images.
    \item \textbf{CUB-200} \cite{CUB}: This dataset consists of images of 200 species of birds with first 100 species for training (5864 examples) and remaining for testing (5924 examples).
    \item \textbf{Cars-196} \cite{Cars196}: It consists of images of cars belonging to 196 models. The first 98 models containing 8054 images are used for training. The remaining 98 models containing 8131 images are used for testing.
\end{itemize}
The MNIST, Fashion-MNIST and CIFAR-10 datasets are widely used benchmarks with sufficiently large number of images for a comparative evaluation of different approaches. The CUB-200 and Cars-196 datasets are well-known for their use in \textit{Fine-Grained Visual Categorization} (FGVC), and have huge intra-class variances and inter-class similarities.

\textbf{Implementation details:} We adapted the network architectures in the MatConvNet tool \cite{MatConvNet}. For MNIST and Fashion datasets, the network for MNIST has been adapted as: Conv1($5\times5,20$) $\to$ max-pool $\to$ Conv2($5\times5,50$) $\to$ max-pool $\to$ Conv3($4\times4,500$) $\to$ ReLU $\to$ FC($500\times128$) $\to$ $l2$ $\to$ $\Mat{L}$($128\times64$). For CIFAR-10, we used the following adapted network: Conv1($5\times5,32$) $\to$ max-pool $\to$ ReLU $\to$ Conv2($5\times5,32$) $\to$ ReLU $\to$ avg-pool $\to$ Conv3($5\times5,64$) $\to$ ReLU $\to$ avg-pool $\to$ Conv4($4\times4,64$) $\to$ ReLU $\to$ $l2$ $\to$ $\Mat{L}$($64\times32$). For our method, we set $\gamma=0.99$ in the affinity propagation step, $k=10$ in the kNN graph, $\alpha=40^\circ$ in (\ref{opt_prob_AGML}), and initial learning rate $10^{-4}$. For MNIST and Fashion datasets, we choose 100 labeled examples (10 per class), while for CIFAR-10, we choose 1000 labeled examples (100 per class). We sample a random subset of 9k unlabeled examples and use it along with the labeled data to mine triplets. For each random subset, we run our method for 10 epochs (with mini-batch size of 100 triplets). In total, we run for a maximum of 50 epochs and choose the best model from across all runs. For MNIST and Fashion, we train upon randomly initialized networks. For CIFAR-10, we observed a better performance by pretraining with labeled examples (by replacing the $\Mat{L}$ \textit{layer} with softmax) for 30 epochs, and then fine-tune using our loss for 50 epochs. For all datasets, the graph has been constructed using the $l2$-normalized representations obtained just before $\Mat{L}$.

For the FGVC task, the GoogLeNet \cite{GoogLeNet} architecture pretrained on ImageNet \cite{ImageNet2015}, has been used as the backbone CNN, using MatConvNet \cite{MatConvNet}. We used the Regional Maximum Activation of Convolutions (R-MAC) \cite{RMAC_ICLR16} right before the average pool layer, aggregated over three input scales ($512$, $512/\sqrt{2}$, $256$). We choose five labeled examples per class. For our method, we set $\gamma=0.99$, $k=50$, $\alpha=45^\circ$ in (\ref{opt_prob_AGML}), and embedding size of 128. We take the entire dataset without partition based sampling and run for a maximum of 200 epochs (mini-batch size of 100 triplets), and choose the best model. Specifically, using our triplet mining strategy, we mined $146600$ triplets for the CUB dataset, and $201350$ triplets for the Cars dataset. In all experiments, we fix a validation dataset by sampling $15\%$ examples from each class of the training data. This validation dataset is used to tune the hyperparameters without taking any feedback from the test data. Note that to learn $\Mat{L}$ with orthogonal constraints we made use of the Manopt \cite{manopt} tool with CGD ($max\_iter=10$, with all other default settings).

\textbf{Compared state-of-the-art baseline approaches:}
We compare our proposed SSDML method against the following baseline techniques:
\begin{itemize}[noitemsep]
    \item \textbf{Deep-LRML}: This is the stochastic extension of the classical LRML method \cite{LRML-b} discussed earlier, by making use of our proposed stochastic Algorithm \ref{alg_AGML}. It follows the \textit{min-max principle} for the labeled data (minimizing distances between similar pairs, while maximizing distances between dissimilar pairs). A Laplacian regularizer is used to capture information from the unlabeled data.
    \item \textbf{Deep-ISDML}: Stochastic extension of the ISDML \cite{ISDML} method. It is similar to LRML, but makes use of densities around an example to adapt the Laplacian regularizer.
    \item \textbf{Deep-SERAPH}: Stochastic extension of the SERAPH \cite{SERAPH-b} method that makes use of the entropy minimization principle.
    \item \textbf{Deep-APLLR}: Stochastic extension of the APLLR \cite{APSSML_ICANN18} method. Makes use of a Log-Likelihood Ratio (LLR) based prior metric. The affinity propagation principle is used to propagate information from the labeled pairs to the unlabeled ones, and adapt the Laplacian (but no triplet mining like ours).
    \item \textbf{Deep-APIT}: Stochastic extension of the APIT \cite{APSSML_ICANN18} method. Makes use of an information-theoretic prior metric. The affinity propagation principle is used to adapt the Laplacian as in APLLR.
    \item \textbf{Exemplar} (TPAMI'16): This method \cite{ExemplarCNN_TPAMI16} attempts to learn the parameters associated with certain elementary transformation operations like translation, scaling, rotation, contrast, and colorization applied on random image patches.
    \item \textbf{Rotation Net} (ICLR'18): This method \cite{rotation_ICLR18} aims to learn representations of data that can accurately capture the information present in an image despite any rotation of the subject.
    \item \textbf{NCE} (CVPR'18): This method \cite{NCE_CVPR18} aims at bringing augmentations of an example together while moving away augmentations from different examples.
    \item \textbf{DeepCluster} (ECCV'18): This method \cite{DeepCluster_ECCV18} aims to jointly cluster metric representations while learning pseudo-labels for data in an end-to-end manner.
    \item \textbf{Synthetic} (AAAI'20): This method \cite{SUML_AAAI20} learns a metric using synthetic constraints generated in an adversarial manner.
    \item \textbf{Triplet} (CVPR'15): This method \cite{FaceNet} learns a metric using a standard triplet loss.
    \item \textbf{Angular} (ICCV'17): This method \cite{Angular_loss} learns a metric using an angular loss on a triplet of examples.
\end{itemize}

\textbf{Observations:}
For comparing the approaches, we first learn a metric with the training data, and using it we obtain the test embeddings. The methods are compared based on their clustering (wrt NMI) and retrieval (wrt Recall@K, K=1,2,4,8) performances on the test embeddings. NMI is defined as the ratio of mutual information and the average entropy of clusters and entropy of actual ground truth class labels. The Recall@K metric gives us the percentage of test examples that have at least one K nearest neighbor from the same class. A higher value of all these metrics indicates a better performance for an approach.

The methods Deep-LRML, Deep-ISDML, Deep-SERAPH, Deep-APLLR and Deep-APIT are \textit{semi-supervised} in nature. They have been chosen as direct counterparts for our proposed SSDML method. These baselines, including ours make use of both labeled (limited in number) and unlabeled examples to learn a metric. As seen in Tables \ref{ssdml_all_sota}-\ref{results_all_cub_cars}, we outperform the baselines on all the datasets that vary in sizes and complexities. In Figure \ref{quali_cars}, we also show some qualitative retrieval comparisons of our method against the baselines on the Cars-196 dataset (successful retrieval is shown in green and failure is shown in red). Our method performs fairly better retrieval than the baselines.

The Exemplar, Rotation Net, NCE, DeepCluster and Synthetic techniques are \textit{unsupervised} in nature, i.e., they learn a metric using only unlabeled data. As shown in Table \ref{cars_vs_unsup}, a better performance of our method in contrast to these approaches demonstrates the benefit of additionally available labeled examples while learning a metric.

Lastly, the Triplet and Angular methods are fully-supervised baselines that only learn from labeled examples. As shown in Table \ref{fgvc_vs_sup}, a better performance of our method in contrast to these approaches demonstrates the benefit of leveraging additional unlabeled examples via affinity propagation and our mining technique. On the other hand, the fully-supervised baselines overfit to the training data because of the availability of only a limited number of labeled examples.

\textbf{Ablation studies:} We now perform a few ablation studies to understand the different components of our method. Table \ref{ablation_quant_orthog} reports the performance of our method for the two cases: i) \textbf{w/o orth}: $\Mat{L}$ is not orthogonal, and ii) \textbf{w/ orth}: $\Mat{L}$ is orthogonal. We observed that with orthogonality, the performance is better. To observe this qualitatively, we plot the tSNE embeddings of the test examples of MNIST and CIFAR-10 in Figure \ref{ablation_alpha}-(Right). We could observe that with orthogonality the embeddings are relatively better separated. Especially, in contrast to the simpler MNIST, orthogonality seems to have a more profound effect on the complex CIFAR-10 dataset. As conjectured, without orthogonality one may learn degenerate embeddings due to a model collapse. Additionally, for the Cars-196 dataset we show the convergence behaviour of our method for the first 100 epochs in Figure \ref{ablation_alpha}-(Left), and sensitivity of our method towards $\alpha$ in Figure \ref{ablation_alpha}-(Middle). The performance is fairly stable in the range $35^\circ - 55^\circ$.


\begin{table}[t]
\centering
\resizebox{0.9\columnwidth}{!}{%
\begin{tabular}{|c|c|ccccc|}
\hline
Dataset                                                                            & Method           & \textbf{NMI}  & \textbf{R@1}  & \textbf{R@2}  & \textbf{R@4}  & \textbf{R@8}  \\ \hline
\multirow{2}{*}{\textbf{MNIST}}                                                    & w/o orth         & 42.9          & 91.5          & 95.3          & 97.3          & 98.6          \\
                                                                                   & \textbf{w/ orth} & \textbf{47.5} & \textbf{93.9} & \textbf{96.6} & \textbf{98.2} & \textbf{98.9} \\ \hline
\multirow{2}{*}{\textbf{\begin{tabular}[c]{@{}c@{}}Fashion-\\ MNIST\end{tabular}}} & w/o orth         & 50.3          & 73.3          & 82.4          & 89.7          & 94.4          \\
                                                                                   & \textbf{w/ orth} & \textbf{52.1} & \textbf{77.6} & \textbf{86.0} & \textbf{91.8} & \textbf{95.6} \\ \hline
\multirow{2}{*}{\textbf{\begin{tabular}[c]{@{}c@{}}CIFAR-\\ 10\end{tabular}}}      & w/o orth         & 19.0          & 37.4          & 54.2          & 69.7          & 83.4          \\
                                                                                   & \textbf{w/ orth} & \textbf{25.3} & \textbf{41.4} & \textbf{57.3} & \textbf{72.6} & \textbf{84.9} \\ \hline
\end{tabular}%
}
\caption{Quantitative comparison of the performance of our method, without (w/o orth) and with orthogonality (w/ orth) on the metric parameters.}
\label{ablation_quant_orthog}
\end{table}

\begin{table}[t]
\centering

        \resizebox{0.6\columnwidth}{!}{%
        \begin{tabular}{|c|cccc|}
        \hline
        Method      & \textbf{R@1}  & \textbf{R@2}  & \textbf{R@4}  & \textbf{R@8}  \\ \hline
        Triplet  &42.6 &55.2 &66.9 &77.6          \\ 
        Angular  &41.4	&54.5	&67.1	&78.4 \\ \hline
        \textbf{Ours}        & \textbf{44.8} & \textbf{56.9} & \textbf{69.1} & \textbf{79.9} \\ \hline
        
        \end{tabular}%
        }
        \caption{Comparison of our method against supervised deep metric learning baselines on CUB.}
        \label{fgvc_vs_sup}
\vspace{-0.4cm}
\end{table}




\textbf{Why our method addresses the limitations of the existing SSDML methods, and performs better ?}
1. \textbf{LRML / ISDML vs Ours:} Both the LRML and ISDML methods define the affinity matrices directly using the distances among the initial representations. If the initial affinities are poor, the learned metric would be poor as well. On the other hand, our affinity matrix is adapted by the affinity propagation principle, while leveraging labeled data information as well. 2. \textbf{APLLR / APIT vs Ours:} Although APLLR and APIT make use of affinity propagation, they do not mine constraints using the enriched affinity matrix information, whereas we do. While their prior metric may be singular and hence poor, our method has no such dependency on a pre-computed metric. 3. \textbf{SERAPH vs Ours:} In contrast to SERAPH, our method has a stronger regularizer by virtue of orthogonality.

\section{Conclusions}
In this paper we revisit and revamp the important problem of Semi-Supervised Distance Metric Learning (SSDML), in the end-to-end deep learning paradigm. We discuss a simple stochastic approach to extend classical SSDML methods to the deep SSDML setting. Owing to the theoretical limitations of the existing SSDML techniques, we propose a method to address these limitations. While the components in our method have been studied earlier, their composition has not been performed before. We show that by using this design composition, the overall approach could outperform existing approaches by the collective use of some of the best practices. In short, following are our major contributions: 1. Extension of classical SSDML approaches for end-to-end stochastic deep learning, with the proposal of a new method, 2. A novel triplet mining strategy leveraging graph-based affinity propagation, and 3. Adaptation of an angular variant of a metric learning loss with orthogonality imposed on the metric parameters to avoid a model collapse.

\section{Acknowledgements}
The work was initiated when UKD was in IIT Madras. We would like to thank the anonymous reviewers for their efforts at reviewing our work and providing valuable comments to polish our paper.

\bibliography{AGML_AAAI21}

\end{document}